\documentclass[10pt, conference, compsocconf]{IEEEtran}
\usepackage{algorithm}
\usepackage{algorithmic}
\usepackage{mathtools}
\usepackage{amssymb}

\begin{document}
\title{Weighted Spectral Cluster Ensemble}


%
\author{\IEEEauthorblockN{Muhammad~Yousefnezhad}
	\IEEEauthorblockA{Department of Computer Science and Technology\\
		 Nanjing University of Aeronautics and Astronautics\\
		Nanjing, China\\
		myousefnezhad@nuaa.edu.cn}
	\and
	\IEEEauthorblockN{Daoqiang~Zhang}
	\IEEEauthorblockA{Department of Computer Science and Technology\\
		Nanjing University of Aeronautics and Astronautics\\
		Nanjing, China\\
		dqzhang@nuaa.edu.cn}
}
\maketitle

\begin{abstract}
Clustering explores meaningful patterns in the non-labeled data sets. Cluster Ensemble Selection (CES) is a new approach, which can combine individual clustering results for increasing the performance of the final results. Although CES can achieve better final results in comparison with individual clustering algorithms and cluster ensemble methods, its performance can be dramatically affected by its consensus diversity metric and thresholding procedure. There are two problems in CES: 1) most of the diversity metrics is based on heuristic Shannon's entropy and 2) estimating threshold values are really hard in practice. The main goal of this paper is proposing a robust approach for solving the above mentioned problems. Accordingly, this paper develops a novel framework for clustering problems, which is called Weighted Spectral Cluster Ensemble (WSCE), by exploiting some concepts from community detection arena and graph based clustering. Under this framework, a new version of spectral clustering, which is called Two Kernels Spectral Clustering, is used for generating graphs based individual clustering results. Further, by using modularity, which is a famous metric in the community detection,  on the transformed graph representation of individual clustering results, our approach provides an effective diversity estimation for individual clustering results. Moreover, this paper introduces a new approach for combining the evaluated individual clustering results without the procedure of thresholding. Experimental study on varied data sets demonstrates that the prosed approach achieves superior performance to state-of-the-art methods.
\end{abstract}

%
\IEEEpeerreviewmaketitle

\section{Introduction}
Clustering, the art of discovering meaningful patterns in the non-labeled data sets, is one of the main tasks in machine learning. Generally, individual clustering algorithms provide different accuracies in a complex data set because they generate the clustering results by optimizing a local or global function instead of natural relations between data points in each data set. \cite{Strehl02, Fred08}. As a novel solution, cluster ensemble which combines the different clustering results was proposed for achieving a better final result \cite{Strehl02}. Cluster Ensemble Selection (CES) is a new solution which combines a selected group of best individual clustering results according to consensus metric(s) from ensemble committee in order to improve the accuracy of final results \cite{Fern08}. The evaluation metric(s), thresholding and selection strategy, and aggregation method are the most important challenges in CES for selecting better partitions of ensemble committee and generating the final result. There are a wide range of ideas for solving mentioned challenges \cite{Fern08,Azimi09,Alizadeh14, Jia12, Yousefnezhad13}. Although these methods can improve performance and robustness of final results, using a wide range of threshold values and employing the entropy based metric are two main weak points of this method. Threshold values are different for each data set in the mentioned methods; and it is really hard to find optimum values in real-world applications. Moreover, most of the real-world data sets do not have logarithm behavior. So, there is no prove that entropy based methods, which estimate the diversity based on the logarithm, were the best choice to evaluate the diversity. This paper proposes a novel methodology for solving clustering problems \textit{without mentioned weak points}. 

As mentioned before, there are four stages in Cluster Ensemble Selection (CES); i.e. generating individual clustering results, evaluating, selecting and combining them as a final clustering result. Although CES can achieve a better result in comparison with individual clustering algorithms and cluster ensemble methods, the accuracy of CES is fully sensitive to the process of thresholding for selecting individual clustering results, and the consensus metric, which is used for diversity or quality estimation of the results. Unfortunately, it is so hard to find the optimum threshold values in practice; and most of the metrics, which were used for diversity or quality estimation, are heuristic; especially they are based on Shannon’s entropy. The main goal of this paper is solving mentioned problems. This paper proposes a new method for estimating the diversity of generated individual clustering results by using a redefined version of modularity, which is based on expected value and it is introduced for the community detection’s applications. Further, this paper introduces a novel approach for combining the evaluated individual clustering results without the process of thresholding. 

Our contribution in this paper can be summarized as follows: \textit{Firstly,} this study proposes a greedy method based on feedback mechanism \cite{Alizadeh15} which employs the idea of bisecting k-means for generating individual results. \textit{After that,} this paper introduces the Two Kernels Spectral  Clustering (TKSC) for generating individual clustering results. This algorithm generates hybrid individual clustering results, which contains Partitional results and Modular results. Same as simple clustering problems, our method generates Partitional results; and also it generates Modular results, which represented by a graph, as a new alternative for evaluating and combining the individual results. \textit{Next,} to satisfy the diversity criterion, this study proposes Normalized Modularity, which is a redefined version of Modularity criterion in community detection \cite{Clauset04}, for evaluating diversity of individual results in the general clustering problems. Unlike most of the diversity metrics which are based on Shannon's entropy, this metric uses Expected Value in probabilistic theory for evaluating individual clustering results and avoids the undesired logarithm \cite{Clauset04,Newman06}. \textit{Lastly,} this paper proposed Weighted Evidence Accumulation Clustering (WEAC) to obtain the final clustering with a weighted combination of all individual results. While the weight of each individual result in WEAC can be estimated with different metrics, the normalized modularity was used in this paper. 

The rest of this paper is organized as follows:  In Section 2, this study first briefly reviews some related works on cluster ensemble selection. Then, it introduces the proposed Weighted Spectral Clustering Ensemble (WSCE) framework in Section 3. Experimental results are reported in Section 4; and finally this paper presents conclusion and pointed out some future works in Section 5.
\section{Related Works}
As an unsupervised method, Clustering  discovers meaningful patterns in the non-labeled data sets. There is a wide range of studies, which try to increase the performance of clustering algorithms. For instance, Zhang et al. introduced a multi-manifold regularized nonnegative matrix factorization framework (MMNMF) which can preserve the locally geometrical structure of the manifolds for multi-view clustering \cite{Zhang14}. Anyway, individual clustering algorithms provide different accuracies in a complex data set because they generate the clustering results by optimizing a local or global function instead of natural relations between data points in each data set \cite{Strehl02,Fred08}. 

Generally, a cluster ensemble has two important steps: Firstly, generating individual clustering results by using different algorithms and changing the number of their partitions. Then, combining the primary results and generating the final ensemble. This step is performed by consensus functions (aggregating mechanism) \cite{Strehl02,Fred05}. 

The idea that not all partitions are suitable for cooperating to generate the final clustering was proposed in CES \cite{Fern08}. Instead of combing all achieved individual results, CES can combine a selected group of best individual results according to consensus metric(s) from the ensemble committee in order to improve the accuracy of final results \cite{Fern08,Alizadeh14,Alizadeh15,Azimi09,Yousefnezhad13}. Fern and Lin developed a method to effectively select individual clustering results for ensemble and the final decision \cite{Fern08}. Azimi et al. proved that diversity maximization is not an effective approach in some real-world applications. They explored that the thresholding procedure must be done based on the complexity and quality of data sets \cite{Azimi09}. Jia et al. proposed SIM for diversity measurement, which works based on the Normalized Mutual Information (NMI) \cite{Jia12}. Romano et al. proposed Standardized Mutual Information (SMI) for evaluating clustering results \cite{Romano14}. 

Yousefnezhad et al. introduced independency metric instead of quality metric for evaluating the process of solving a problem in the CES \cite{Yousefnezhad13}. Alizadeh et al. have concluded the disadvantages of NMI as a symmetric criterion. They used the APMM\footnote{Alizadeh-Parvin-Moshki-Minaei} and Maximum (MAX) metrics to measure diversity and stability, respectively, and suggested a new method for building a co-association matrix from a subset of base cluster results \cite{Alizadeh14,Alizadeh15}. Alizadeh et al. introduced Wisdom of Crowds Cluster Ensemble (WOCCE), which is a novel method base on a theory in social science \cite{Alizadeh15}. Although, this method can generate high performance and more stable results in comparison with other CES methods, using a wide range of thresholds and employing different types of clustering algorithms for generating individual results are two main problems in this method. Alizadeh et al. used A3, which is based on Shannon's entropy, for diversity evaluation; and Basic Parameter Independency (BPI), which uses initialized values of individual clustering algorithms such as random seeds in the first iterative of k-means, for independency evaluation. In addition, they introduced the feedback mechanism for generating the high-quality results \cite{Alizadeh15}.

As a graph based clustering methods, spectral clustering generates high-performance results when it is applied to different applications; i.e. from image segmentation to community detection arena. Kuo et al. introduced a new method for automating the process of Laplacian creation in the medical applications; especially for fMRI segmentation where this method used standard Laplacians perform poorly \cite{Kuo14}. Chen et al. proposed a clustering algorithm which is based graph clustering and optimizing an appropriate weighted objective, where larger weights are given to observations (edge or no-edge between a pair of nodes) with lower uncertainty \cite{Chen14}. Gao et al. introduced a graph-based consensus maximization (BGCM) method for combining multiple supervised and unsupervised models. This method consolidated a classification solution by maximizing the consensus among both supervised predictions and unsupervised constraints \cite{Gao13}. 
\section{The proposed method}
Given a set of high-dimensional data examples $\hat{X} = \{\hat{x}_{1},\hat{x}_{2},\ldots,\hat{x}_{n}\}$. The simple average of $\hat{X}$ can be denoted as follows:
\begin{equation}
\bar{X} = \frac{1}{n}\sum_{i=1}^{n}\hat{x}_{i}
\end{equation}
where $n$ is the number of instances in the $\hat{X}$; and $\hat{x}_{i}$ denotes the $i-th$ instance of the data points. At the beginning, this paper minimized the correlation between features. So, it denotes $X$ as follows:
\begin{equation}
X = \hat{X} - \bar{X} = \{(\hat{x}_{1} - \bar{x}_{1}), (\hat{x}_{2} - \bar{x}_{2}), \ldots, (\hat{x}_{n} - \bar{x}_{n})\}
\end{equation}
where $\hat{X}$ is the data points, and $\bar{X}$ denotes simple average of $\hat{X}$, which calculated by (1). It's clear that $X$ is zero-mean. In other words, the excepted value of $X$ is zero as follows:
\begin{equation}
\mathbb{E}\{X\}=0
\end{equation}
Now, this paper maps $Q: X\in \mathbb{R}^{m\times n} \to Y\in \mathbb{R}^{m\times n}$, where $m$, $n$ denote the number of features and data points, respectively. This mapping just minimizes the correlation between features. This problem can be reformulate as follows:
\begin{equation}
Y = {Q}^{T}X
\end{equation}
If the correlation (covariance) of $X$ is considered $R = \mathbb{E}\{X{X}^{T}\} = \frac{1}{n}\sum_{i = 1}^{n}{x}_{i}{x}^{T}_{i} $, then the correlation of $Y$ will be defined as follows:
\begin{equation}
\begin{multlined}
\mathbb{E}\{Y{Y}^{T}\} = \mathbb{E}\{({Q}^{T}X){({Q}^{T}X)}^{T}\} =\\ 
\mathbb{E}\{{Q}^{T}X{X}^{T}Q\} = {Q}^{T}\mathbb{E}\{X{X}^{T}\}Q = {Q}^{T}RQ
\end{multlined}
\end{equation}
Based on above definition, the expected value of $j-th$ feature of $X$ denotes as follows:
\begin{equation}
\mathbb{E}\{{Y}_{j}{Y}^{T}_{j}\} = {q}^{T}R{q}
\end{equation}
where $q$ denotes the $j-th$ index of the $Q$. In other words, our correlation problem is changed to a variance probe. Now, maximizing the $q$ based on the variance of $X$ will be omitted the correlation between features. Since the scale of data after mapping must be same, we assume following equation:   
\begin{equation}
\|q\| = 1
\end{equation}
For maximizing the (6), which is denoted by $\Psi(q)$, our problem will be reformulated as follows:
\begin{equation}
\begin{multlined}
max[\Psi(q) = {q}^{T}Rq] \Rightarrow\\
\frac{\partial\Psi(q)}{\partial q} = 0 \Rightarrow\\
\Psi(q + \delta q) = \Psi(q) \Rightarrow\\
{(q + \delta q)}^{T}R(q + \delta q) = {q}^{T}R{q}
\end{multlined}
\end{equation}
where the symbol $\delta q$ is an abbreviation for `a small change in q'. We consider $(\delta q)^{T}\delta q \approx 0,$ so the above definition denotes as follows:
\begin{equation}
(\delta q)^{T}Rq = 0
\end{equation}
Based on (7) and (8), we can assume as follows:
\begin{equation}
\|\delta q - q\| = \|q\| = 1 \Rightarrow (\delta q)^{T}q = 0
\end{equation}
Now, this paper defines following equation by using (9) and (10):
\begin{equation}
\begin{multlined}
(\delta q)^{T}Rq - \lambda(\delta q)^{T}q = 0 \Rightarrow\\
(\delta q)^{T}[Rq - \lambda q] = 0
\end{multlined}
\end{equation}
where $\lambda \in \mathbb{R}$ is a constant. Since $(\delta q)^{T} \neq 0,$ the following equation must be satisfy for minimizing correlation between features:
\begin{equation}
Rq = q \lambda
\end{equation}
where R and $\lambda$ denotes the eigenvectors and eigenvalues, respectively. For all features of $X$ the above equation will be denoted as follows:
\begin{equation}
RQ = Q \Lambda
\end{equation}
which is called eigenstructure equation. In above equation, $\Lambda$ is a diagonal matrix. Based on (7), we can define following equation:
\begin{equation}
\|q\|^{2} = 1 \Rightarrow {Q}^{T}Q = \mathbb{I}
\end{equation}
where $\mathbb{I}$ is identity matrix. Following equation denotes based on (13) and (14):
\begin{equation}
\begin{multlined}
RQ = Q \Lambda \Rightarrow\\
RQ{Q}^{T} = Q \Lambda{Q}^{T} \Rightarrow\\
R\mathbb{I} = Q \Lambda{Q}^{T} \Rightarrow\\
R = {Q}^{T}\Lambda Q \Rightarrow\\
R = \sum_{j = 1}^{m} \lambda_{j} {q}_{i} {q}^{T}_{j}
\end{multlined}
\end{equation}
where $m$ denotes number of features in data $X$. Now, consider that $R$ is a descending order based on $\Lambda$ values. For an optional feature selection we can define the following equation instead of (15):
\begin{equation}
R = \sum_{j = 1}^{d} \lambda_{j} {q}_{i} {q}^{T}_{j}
\end{equation}
where $d < m$ is the number of features, which must be selected for generating results. Algorithm 1 shows the mapping function, which can minimized correlation of data set based on above definitions. 

\begin{algorithm}
	\caption{The Mapping Function}
	\label{alg:Mapping}
	\begin{algorithmic}
		\STATE {\bfseries Input:} Data set $\hat{X} \in \mathbb{R}^{m \times n}  = \{\hat{x}_{1},\hat{x}_{2},\ldots,\hat{x}_{n}\}$,\\
		\quad $d$ as number of features: \\ 
		\quad \emph{$d = 0$ is considered for deactivating the feature selection}
		\STATE {\bfseries Output:} Mapped data set $Y$
		\STATE {\bfseries Method:}\\
		\quad1. Calculating simple average $\bar{X}$ by using (1).\\
		\quad2. Calculating $X$ by using (2).\\
		\quad3. Calculating $R = \mathbb{E}\{X{X}^{T}\} = \frac{1}{n}\sum_{i = 1}^{n}{x}_{i}{x}^{T}_{i}$.\\
		\quad4. Calculating $\Lambda$ and $Q$ as eigenvalues/vectors of $R$.\\
		\quad5. Sorting $Q$ based on descending values of $\lambda$. \\
		\quad6. \textbf{if} d is not zero ($d \neq 0$)\\
		\quad\quad \textbf{then} selecting $[1, d]$ features of $Q$, and sorting as ${Q}_{d}$,\\ 
		\quad\quad \textbf{else} ${Q}_{d} = Q$, $d = m$.\\ 
		\quad\quad \textbf{end if}\\
		\quad7. Return $Y = {Q}_{d}^{T}X$.\\ 
	\end{algorithmic}
\end{algorithm}
For generating individual clustering results, the proposed method partitions $Y$ into ${C}^{l}$ clusters, where k denotes number of clusters in the individual results, and ${C}^{l}$ is $l-th$ individual result in the reference set. This paper uses the range of $\mathit{l \in [2,k+2]}$ for generating individual results, where $k$ is the number of clusters in the final result. This is the same as bisect k-means algorithms but instead of applying the algorithm on generated results in each iterative, our proposed method stores this result on the ensemble committee; and then evaluates and combines these results. In other words, the reference set denotes $\zeta \in \mathbb{R}^{n \times [2, k+1]} = \{{C}^{l}\} = \{{C}^{2}, \dots, {C}^{k+2}\}$. 

Like other spectral methods, this paper calculates the non-symmetric distances (adjacency) matrix of $Y$, which is denoted by $A$ \cite{ Ng01,Yan09}. In the rest of this paper, our proposed method will be applied to the matrix $A$ for each individual clustering results.  Moreover, this paper uses (17) as transform function for converting distances matrix $A$ to similarity matrix $S$. This transformation can optimize the memory usage \cite{Ng01,Yan09}. 
\begin{equation}
\label{Convert Equation}
{S}_{i,j}= \left \{
\begin{array}{l l}
exp\left(\frac{-{\|{y}_{i}-{y}_{j} \|}_{2}}{{\phi}^{2}} \right) &  \quad\textit{if } i\neq j \\ 
0 & \quad\textit{if } i=j
\end{array}
\right.
\end{equation}
where ${y}_{n}$ denotes the $n-th$ data point and ${\|{y}_{i}-{y}_{j} \|}_{2}$ will be calculated by Euclidean distance. The scaling parameter $\phi$ controls how rapidly affinity ${S}_{i,j}$ falls off with the distance  between the data points. This paper uses Ng et al. method for estimating this value automatically (count non-zero values in each columns of distance matrix $A$) \cite{Ng01,Yan09}.

This paper introduces Two Kernels Spectral Clustering (TKSC) algorithm, which can generate all individual results ($\zeta$). Unlike normal clustering algorithms, which just generate a partition as the result, the TKSC algorithm generates two independent consequences, which are called Partitional result and Modular result, for each of the individual clustering results by using two kernels (${C}^{l} = \{{P}^{l}, M\} $). Partitional result (${P}^{l}$) is a partitioning of data points same as the result of other clustering methods; and Modular result ($M$) is a network of data points, which can be represented by a graph. This paper uses Modular result as a reference for evaluating the diversity of generated partition by using community detection methods \cite{Clauset04, Newman06}. Furthermore, kernel in the TKSC refers to Laplacian equation in spectral methods because it transforms data points in new environment, especially linear environment for non-linear data sets.

\textbf{Partitional Kernel:} This paper uses following equation for generating Partitional result:
\begin{equation}
\label{Partitional Kernel}
{L}_{P}=\mathbb{I}-{D}^{1/2}S{D}^{1/2} 
\end{equation}
where $\mathbb{I}$ is the identity matrix \cite{Ng01};  $D$ is the diagonal matrix of $S$ ($D = diag(S)$); and $S$ will be calculated by (17). As shows in follows, the eigendecomposition is performed for calculating eigenvectors of ${L}_{P}$:
\begin{equation}
\label{Eigen decomposition}
V = eigens({L}_{P})
\end{equation}
where the matrix $V$ is the eigenvectors of Partitional Kernel. The coefficient $W$ will be defined for normalizing the matrix $V$:
\begin{equation}
\label{W Coefficinet}
{W}_{i} = {\left(\sum_{i=1}^{n}{V}_{i1} \times {V}_{i2}\right)}^{\frac{1}{2}} + \epsilon
\end{equation}
where ${V}_{ij}$ shows the i-th row and j-th column of the matrix $V$; and $\epsilon$ is used for omitting the effect of zeros in the matrix $W$. This paper uses $\epsilon={10}^{-20}$ for generating the experimental results. Also, $n$ denotes the number of instances in the data set ($W \in \mathbb{R}^{n}$). The normalized matrix of eigenvectors will be calculated as follows:
\begin{equation}
\label{U_ij, Normalized Basic Results }
{U}_{ij} = {V}_{ij} \times {W}_{i}
\end{equation}
where ${U}_{ij}$ and ${V}_{ij}$ denote the i-th row and j-th column of these matrices; and ${W}_{i}$ is the i-th row of the matrix $W$ which is used for normalization. The Partitional result of TKSC will be calculated by applying  the simple k-means \cite{Alizadeh15} on the matrix $U$ as follows:
\begin{equation}
\label{Partitional Result}
{P}^{l} = kmeans(U, l)
\end{equation}    
where $K$ is the number of classes in individual results; and U will be calculated by (21).

\textbf{Modular Kernel:} This paper uses following equation for generating Modular result:
\begin{equation}
\label{Modular Kernel}
{L}_{M}=D-S 
\end{equation}
where $D$ is the diagonal matrix of $S$ ($D = diag(S)$); and $S$ will be calculated by (17). This paper considers the normalized matrix of ${L}_{M}$ an adjacency matrix of graph representation of individual result as follows:  
\begin{equation}
\label{Modular Result}
M = \frac{1}{max({L}_{M})}{L}_{M}
\end{equation}
where ${L}_{M}$ is calculated by (23), and the function $max$ finds the biggest value in the matrix ${L}_{M}$. Further, all values in the matrix $M$, which is called Modular result, are between zero and one. Algorithm 2 shows the pseudo code of the TKSC method.
\begin{algorithm}
	\caption{Two Kernels Spectral  Clustering (TKSC)}
	\label{alg:TKSC}
	\begin{algorithmic}
		\STATE {\bfseries Input:} Distance matrix $A$, Number of clusters $l$
		\STATE {\bfseries Output:} Partitional result ${P}^{l}$, Modular result $M$
		\STATE {\bfseries Method:}\\
		\quad1. Generate similarity matrix $S$ by using $A$ on (17).\\
		\quad2. Generate diagonal matrix $D$ by using $S$.\\
		\quad3. Generate ${L}_{P}$ by applying $S$ and $D$ on (18).\\
		\quad4. Generate ${L}_{M}$ by using $S$ and $D$ on (23).\\
		\quad5. Generate the matrix $V$ as eigenvectors of ${L}_{p}$.\\
		\quad6. Generate $U$ as normalized $V$ by using (20) and (21).\\
		\quad7. Generate $M$ by applying ${L}_{M}$ on (24).\\
		\quad8. ${P}^{l} = kmeans(U, l)$\\  
		\quad9. Return ${P}^{l}$ and $M$
	\end{algorithmic}
\end{algorithm}
Tracing errors can control similarity and repetition of specific answers in clustering problems. There is a wide range of metrics, which are based on Shannon's entropy\cite{Alizadeh15,Alizadeh14}, for evaluating the diversity of individual results in the CES methods, such as MI \cite{Strehl02}, NMI \cite{Fred05}, APMM \cite{Alizadeh14}, MAX \cite{Alizadeh15}, and SMI \cite{Romano14}. Shannon's entropy uses the logarithm of probability of individual results for evaluating the diversity but there is no mathematical prove that all real-world data sets have logarithmic behavior. In community detection arena \cite{Clauset04,Newman06}, Modularity, which is based on Expected Value, was proposed for solving this problem. Recently, many papers proved that modularity \cite{Clauset04,Newman06} can estimate the diversity on graph data sets better than entropy based methods. Unfortunately, modularity can measure the diversity only for graph data \cite{Clauset04}. This paper proposes TKSC, which can generate a graph based result, called Modular result, for any types of data sets in real-world application. Since modularity was defined for community detection arena, this paper introduces a redefined version of modularity metric for general clustering problems, which is called Normalized Modularity ($NM$). It is used for evaluating the diversity of the individual results based on Modular result of the TKSC as follows:    
\begin{equation}
\label{Normalized Modularity}
NM({P}^{l}, M) = \frac{1}{2} + \frac{1}{4z}\sum_{ij}^{}\left[{\Gamma}_{ij} - \frac{{\sigma}_{i}{\sigma}_{j}}{2z}\right]\Theta\left({c}_{i}, {c}_{j}\right)
\end{equation}
where ${P}^{l}$ and $M$ are calculated by (22) and (24), respectively; $z$ is sum of all cells in the matrix $M$ ($z=\sum_{M}^{}{M}_{ij}$); and ${c}_{i}$ and ${c}_{j}$ are the cluster's numbers of the i-th and j-th instances in the Partitional result ${P}^{l}$.  Also, ${\sigma}_{i}$ and ${\sigma}_{j}$ show the degree of i-th and j-th nodes in the graph of matrix $M$ (How many rows contains non-zero value in the columns $i$ or $j$). In addition ${\Gamma}_{ij}$ and $\Theta\left({c}_{i}, {c}_{j}\right)$ will be calculated as follows:
\begin{equation}
\label{Ga_ij}
{\Gamma}_{ij}= \left \{
\begin{array}{l l}
0 &  \quad\textit{if } {M}_{ij}=0 \\ 
1 & \quad\textit{Otherwise}
\end{array}
\right.
\end{equation}
\begin{equation}
\label{Thet_ij}
\Theta\left({c}_{i}, {c}_{j}\right)= \left \{
\begin{array}{l l}
1 &  \quad\textit{if } {c}_{i}={c}_{j} \\ 
0 & \quad\textit{Otherwise}
\end{array}
\right.
\end{equation}
This diversity evaluation is $0\le NM\le 1$. In the rest of this section, we describe how $NM$ will be used for evaluating individual clustering results. Thresholding is used for selecting the evaluated individual results in the CES. Then \textit{co-association} matrix is generated by using consensus function on the selected results. Lastly, the final result  is generated by applying linkage methods on the co-association matrix. These methods generate the Dendrogram and cut it based on the number of clusters in the result \cite{Fred05,Alizadeh15}. In recent years, many papers have used EAC as a high-performance consensus function for combining individual results \cite{Fred05,Alizadeh14,Alizadeh15,Azimi09,Fern08}. EAC uses the number of clusters shared by objects over the number of partitions in which each selected pair of objects is simultaneously presented for generating each cell of the co-association matrix.
\begin{figure}[!t]
\centering
\includegraphics[width=2.5in]{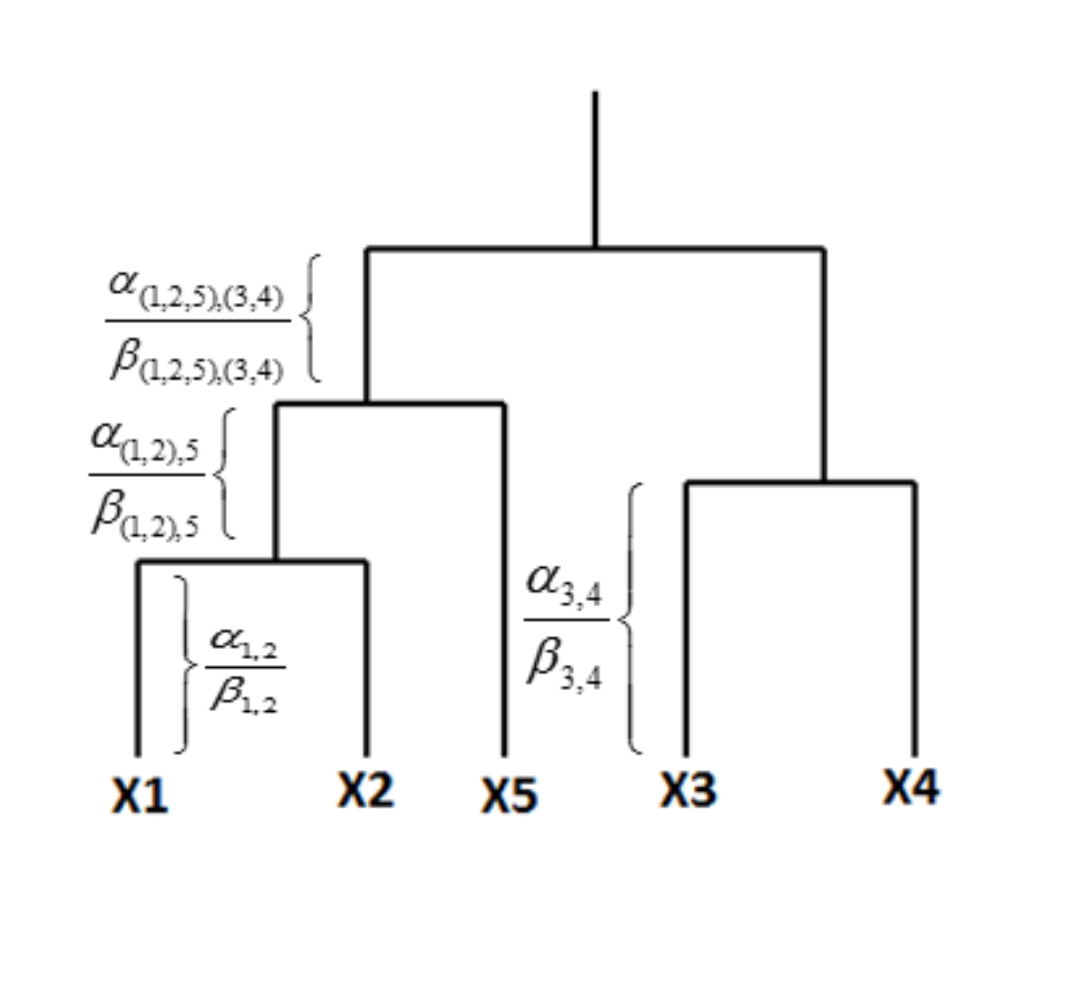}
\caption{
	In the traditional EAC, the ${\alpha}_{(i,j)}$ represents the number of clusters shared by objects with indices (i, j); and ${\beta}_{(i,j)}$ is the number of partitions in which this pair of instances (i and j) is simultaneously presented. This method assumes the weights of all individual clustering results (${\alpha}_{(i,j)}$) are the same. This paper proposes Weighted EAC for optimizing this method by using a weight for each individual clustering results instead of just counting their shared clusters. While the weight can have different definitions in the other applications, this paper uses average of Normalized Modularity (NM) of two algorithms as the weight in the WEAC ($\bar{{\alpha}}_{(i,j)}= \sum_{\alpha\left(i,j\right)}{\rho}_{i,j}$).}
\label{Fig1}
	\vskip -0.2in
\end{figure}
Figure 1 illustrates the effect of the EAC equation ($c\left(i,j\right)=\frac{\alpha\left(i,j\right)}{\beta\left(i,j\right)}$) on the shape of Dendrogram. Where ${\alpha}_{(i,j)}$ represents the number of clusters shared by objects with indices (i, j); and ${\beta}_{(i,j)}$ is the number of partitions in which this pair of instances (i and j) is simultaneously presented. As a matter of fact; EAC considers that the weights of all algorithms’ results are the same. Instead of counting these indices, this paper uses following equation, which is called Weighted EAC (WEAC), for generating the co-association matrix.
\begin{equation}
\label{WEAC}
c\left(i,j\right)=\frac{\sum_{\alpha\left(i,j\right)}{\rho}_{i,j}}{\beta\left(i,j\right)}
\end{equation} 
where $\alpha\left(i,j\right)$ and $\beta\left(i,j\right)$ are same as the EAC equation; Also, ${\rho}_{i,j}$ is the weight of combining the instances. Although this weight can have different definitions in the other applications, this paper uses average of Normalized Modularity of two algorithms as follows for combining individual results:  
\begin{equation}
\label{AvrageOfNormalized Modularity}
{\rho}_{ij}=\frac{1}{2}({NM}_{i} + {NM}_{j})
\end{equation}   
where ${NM}_{i}$ and ${NM}_{j}$ illustrates the Normalized Modularity of the algorithms, which generate the results for indices $i$ and $j$. In other words, as a new mechanism, this paper generates the effective results when both algorithms have high NM values; and also the effects of individual results are near of zero when the both algorithms have small values in the NM metric. As a result, this paper just omits the effect of low quality individual results by using mentioned mechanism instead of selecting them by thresholding procedures. Further, the final co-association matrix, which is a symmetric matrix, will be generated by (28) as follows:
\begin{equation}
	\label{co-association matrix}
	\xi = WEAC(\zeta) = \left( 
	\begin{array}{cccc}
		c(1,1) & c(1,2) & \dots & c(1,n) \\
		c(2,1) & c(2,2) & \dots & c(2,n) \\
		\vdots & \vdots & \vdots & \vdots \\
		c(i,1) & c(i,2) & c(i,j) & c(i,n) \\
		\vdots & \vdots & \vdots & \vdots \\
		c(n,1) & c(n,2) & \dots & c(n,n) \\
		\end{array} \right)
\end{equation}   
where $n$ is the number of data points; and $c(i,j)$ denotes the final aggregation for $i-th$ and $j-th$ instances. Algorithm 3 illustrates the pseudo code of the proposed method. In this algorithm, $\hat{X}$ is the data set; k is the number of clusters in the final result; ${P}_{f}$ is the final result partition. The distances are also measured by an Euclidean metric. The TKSC function builds the partitions and modules of individual results; and NM function evaluates these results by using (25). Then, the evaluated results will be added to reference set. The WEAC function generates the co-association matrix, according to (28), by using the Normalized Modularity values and Partitional results. The Average-Linkage function creates the final ensemble according to the average linkage method \cite{Alizadeh14,Alizadeh15}. 
\begin{algorithm}
	\caption{The Weighted Spectral Cluster Ensemble}
	\label{alg:WSCE}
	\begin{algorithmic}
		\STATE {\bfseries Input:}\\
		\quad Data points $\bar{X}$,\\ 
		\quad Number of clusters $k$,\\
		\quad Number of features d.\\	
		\STATE {\bfseries Output:} final result ${P}_{f}$\\
		\STATE {\bfseries Method:} \\
		1.\quad Generate Y by using $\bar{X}$ and d on Algorithm 1.\\
		2.\quad Generate matrix $A$ by using $Y$ based on \cite{Ng01}.\\
		3.\quad \textbf{for} $l=2$ \textbf{to} $k+2$ \textbf{do}\\
		4.\quad\quad$\left[{P}^{l}, M\right]=TKSC\left(A, l\right)$ by using Algorithm 2.\\ 
		5.\quad\quad$Q = NM({P}^{l}, M)$ based on (25)\\
		6.\quad\quad Add $\left[P, Q\right]$ to \textit{$\zeta$} as the reference set.\\
		7.\quad \textbf{end for}\\
		8.\quad Generate co-association matrix $\xi =  WEAC(\zeta)$\\
		9.\quad ${P}_{f} = Average-Linkage(\xi)$
	\end{algorithmic}
\end{algorithm}
\section{Experiments}
The empirical studies will be presented in this section. The unsupervised methods are used to find meaningful patterns in non-labeled datasets such as web documents, etc. in real world application. Since the real dataset doesn’t have class labels, there is no direct evaluation method for estimating the performance in unsupervised methods. Like many pervious researches \cite{Fred05,Fern08,Alizadeh14,Alizadeh15,Yousefnezhad13}, this paper compares the performance of its proposed method with other individual clustering methods and cluster ensemble (selection) methods by using standard datasets and their real classes. Although this evaluation cannot guarantee that the proposed method generate high performance for all datasets in comparison with other methods, it can be considered as an example for analyzing the probability of predicting good results in the WSCE.

The results of the proposed method are compared with individual algorithms k-means \cite{Alizadeh15} and Maximum Likelihood Estimator (MLE) \cite{Chen14}, as well as APMM \cite{Alizadeh14}, WOCCE \cite{Alizadeh15}, SMI \cite{Romano14}, and BGCM \cite{Gao13} which are state-of-the-art cluster ensemble (selection) methods. This paper reported the empirical results of k-means algorithm as one of the classical clustering methods. Furthermore, as a new alternative in the clustering methods, the empirical results of the proposed method are compared with the MLE, SMI, and BGCM methods. Also, this paper uses the unsupervised version of BGCM method (with the null set of supervision information). For representing the effect of Uniformity on the performance of the final results, it compares with two state-of-the-art metrics in diversity evaluation (APMM and SMI). The last but not least, the experimental results of this paper are compared with the WOCCE as another method in the CES, which uses the independency estimation. All of these algorithms are implemented in the MATLAB R2015a (8.5) by authors\footnote{The proposed method is available http://sourceforge.net/projects/myousefnezhad/files/WSCE/} in order to generate experimental results. All results are reported by averaging the results of 10 independent runs of the algorithms which are used in the experiment. Also, the number of individual clustering results in the reference set of the ensemble is set as 20 for all of mentioned algorithms in all of experiments on a PC with certain specifications\footnote{Apple Mac Book Pro, CPU = Intel Core i7 (4*2.4 GHz), RAM = 8GB, OS = OS X 10.10}.
\begin{table}
	\caption{The standard data sets}
	\label{Tbl1: Data Set}
	\begin{center}
		\begin{small}
			\begin{tabular}{lccc}
				\hline
				Data Set & Instances & Features & Class \\
				\hline
				20 Newsgroups & 26214 & 18864 & 20 \\
				ADNI-MRI-C1  & 202  & 93 & 3 \\
				ADNI-MRI-C2  & 202  & 93 & 4 \\
				ADNI-PET-C1  & 202  & 93 & 3 \\
				ADNI-PET-C2  & 202  & 93 & 4 \\
				ADNI-FUL-C1  & 202 & 186 & 3 \\
				ADNI-FUL-C2  & 202 & 186 & 4 \\		
				Arcene          &  900 & 10000 & 2  \\
				Bala. Scale    &  625 & 4 & 3 \\
				Brea. Cancer &  286 & 9 &  2 \\
				Bupa             & 345 & 6 & 2 \\ 
				CNAE-9         & 1080 & 857 & 9 \\
				Galaxy           & 323 & 4 & 7 \\
				Glass             &  214 & 10 & 6 \\
				Half Ring        &  400 & 2 & 2 \\
				Ionosphere    & 351 & 34 & 2 \\
				Iris                 & 150 & 4 & 3 \\
				Optdigit          & 5620 & 62 & 10 \\
				Pendigits       & 10992 & 16 & 10 \\
				Reuters-21578 & 9108 & 5 & 10\\
				SA Hart         & 462 & 9 & 2 \\
				Sonar            & 208 & 60 & 2 \\
				Statlog           & 6435 & 36 & 7 \\
				USPS             & 9298 & 256 &  10 \\
				Wine              & 178 & 13 & 2 \\
				Yeast              & 1484 & 8 & 10 \\
				\hline
			\end{tabular}
		\end{small}
	\end{center}
	\vskip -0.1in
\end{table}
\subsection{Data Sets}
This paper uses three different groups of data sets for generating experimental results; i.e. image based data sets, document based data sets and others. Table I illustrates the properties of these data sets. This paper uses the USPS digits data set, which is a collection of $16\times16$ gray-scale images of natural handwritten digits and is available from \cite{USPS}. Furthermore, this paper uses Alzheimer's Diseases Neuroimaging Initiative (ADNI) data set for $202$ subjects as another image based real-world data set. This data set contains MRI and PET images from human Brian in two categories (which are shown by C1 and C2 in the Table I and II) for recognizing the Alzheimer diseases. In the first category, this data set partitions subjects to three groups of Health Control (HC), Mild Cognitive Impairment (MCI), and Alzheimer's Diseases (AD). In the second category, there are four groups because the MCI will be partitioned to high and low risk groups (HMCI/LMCI). This paper uses all possible forms of this data set by using only MRI features, only PET features and all of MRI and PET features (FUL) in each of two categorize. More information about ADNI-202 is available in \cite{Zu14}. As a document based data set, the 20 Newsgroups is a collection of approximately 20,000 newsgroup documents, partitioned (nearly) evenly across 20 different newsgroups. Some of the newsgroups are very closely related to each other, while others are highly unrelated. It has become a popular data set for experiments in text applications of machine learning techniques, such as text classification and text clustering. Moreover, the Reuters-21578 is one of the most widely used test collections for text classification research. This data set was collected and labeled by Carnegie Group, Inc. and Reuters, Ltd. We use the 10 largest classes of this data set. The rest of standard data sets are from UCI \cite{UCI}. The chosen data sets have diversity in their numbers of clusters, features, and samples. Further, their features are normalized to a mean of 0 and variance of 1, i.e. N (0, 1).
\begin{table*}[t]
	\caption{The performance of clustering algorithms. Further, the optional feature selection is not used for the proposed method ($d = 0$).  }
	\label{tbl: Expermintal Results}
	\vskip 0.15in
	\begin{center}
		\begin{small}
			\begin{tabular}{lccccccc}
				\hline
				Data Sets & Spectral & MLE & APMM & WOCCE & SMI & BGCM & WSCE \\
				\hline
				20 Newsgroups & 14.31$\pm$2.14 & 21.89$\pm$1.02 & 28.03$\pm$0.87 & 32.62$\pm$0.52 & 29.14$\pm$0.91 & 40.61$\pm$0.83 & \textbf{52.06$\pm$0.17}  \\
				ADNI-MRI-C1 & 39.24$\pm$0.21 & 39.84$\pm$0.42 & 48.01$\pm$0.56 & 48.82$\pm$0.37 & \textbf{50.69$\pm$0.69} & 45.54$\pm$0.99 & 49.53$\pm$0.19\\
				ADNI-MRI-C2 & 32.72$\pm$0.98 & 26.32$\pm$0.67 & 39.93$\pm$0.29 & 40.22$\pm$0.44 & 38.32$\pm$0.41 & \textbf{42.62$\pm$1.04} & 41.14$\pm$0.71\\
				ADNI-PET-C1 & 43.71$\pm$0.52 & 37.96$\pm$0.87 & 48.37$\pm$0.82 & 49.19$\pm$0.26 & 49.45$\pm$0.62 & 42.1$\pm$0.78 & \textbf{52.05$\pm$0.37}\\
				ADNI-PET-C2 & 37.27$\pm$0.23 & 37.91$\pm$0.83 & 38.53$\pm$0.17 & 39.43$\pm$0.79 & 41.76$\pm$0.47 & 39.1$\pm$1.2 & \textbf{43.11$\pm$0.42}\\
				ADNI-FUL-C1 & 42.63$\pm$0.63 & 42.62$\pm$0.58 & 47.22$\pm$0.93 & 48.82$\pm$0.41 & 47.93$\pm$0.83 & 48.56$\pm$1.26 & \textbf{49.06$\pm$0.36}\\
				ADNI-FUL-C2 & 39.51$\pm$1.19 & 41.06$\pm$0.17 & 50.09$\pm$0.35 & 49.39$\pm$0.63 & 49.16$\pm$0.26 & 46.91$\pm$0.42 & \textbf{50.11$\pm$0.09}\\					
				Arcene         & 58.31$\pm$1.22 & 64.19$\pm$0.498 & 66.28$\pm$0.216 & 65.16$\pm$0.32 & 67.14$\pm$0.93 & 64.23$\pm$0.28 & \textbf{73.34$\pm$0.92}\\
				Bala. Scale  & 49.21$\pm$0.87 & 52.76$\pm$0.12 & 52.65$\pm$0.63 & 54.88$\pm$0.61 & 59.98$\pm$0.812 & 59.62$\pm$0.32 & \textbf{61.64$\pm$0.12}\\
				Breast Can. & 94.88$\pm$1.14 & 82.65$\pm$0.342 & 96.04$\pm$0.88 & 96.92$\pm$0.77 & 80.87$\pm$0.652 & 99.12$\pm$0.62 & \textbf{99.21$\pm$0.43}\\
				Bupa           & 56.72$\pm$1.18 & 53.98$\pm$0.274 & 55.07$\pm$0.28 & 57.02$\pm$0.46 & 58.49$\pm$0.21 & 53.17$\pm$0.21 & \textbf{60.93$\pm$0.09}\\
				CNAE-9      & 65.32$\pm$0.43 & 77.72$\pm$0.591 & 77.42$\pm$0.792 & 79.2$\pm$0.579 & 74.25$\pm$0.614 & 80.12$\pm$0.459 & \textbf{88.42$\pm$0.02}\\
				Galaxy        & 31.24$\pm$0.67 & 34.25$\pm$0.872 & 33.72$\pm$0.36 & 35.88$\pm$0.81 & 35.21$\pm$0.413 & 36.91$\pm$0.17 & \textbf{39.89$\pm$0.82}\\
				Glass          & 45.78$\pm$0.87 & 50.32$\pm$0.42 & 47.19$\pm$0.21 & 51.82$\pm$0.92 & 54.19$\pm$0.144 & 53.66$\pm$0.98 & \textbf{55.19$\pm$0.51}\\
				Half Ring    & 80.61$\pm$1.15 & 73.91$\pm$0.762 & 80$\pm$0.42 & 87.2$\pm$0.14 & 71.19$\pm$0.621 & 98.37$\pm$0.59 & \textbf{99.92$\pm$0.08}\\
				Ionosphere & 69.71$\pm$0.67 & 25.67$\pm$0.53 & 70.94$\pm$0.13 & 70.52$\pm$0.132 & 70.87$\pm$0.226 & 73.67$\pm$0.341 & \textbf{76.25$\pm$0.28}\\
				Iris             & 83.45$\pm$0.82 & 89.02$\pm$0.61 & 74.11$\pm$0.25 & 92$\pm$0.59 & 93.79$\pm$0.21 & \textbf{97.29$\pm$0.09} & 96.53$\pm$0.32\\
				Optdigit      & 54.19$\pm$0.45 & 73.81$\pm$0.69 & 77.1$\pm$0.841 & 77.16$\pm$0.21 & 80.21$\pm$0.79 & 71.56$\pm$0.692 & \textbf{82.82$\pm$0.33}\\
				Pendigits    & 53.94$\pm$0.25 & 59.36$\pm$0.31 & 47.4$\pm$0.699 & 58.68$\pm$0.18 & 63.74$\pm$0.37 & 63.13$\pm$0.42 & \textbf{65.02$\pm$0.91}\\
				Reuters-21578 & 48.78$\pm$3.19 & 52.58$\pm$1.92 & 65.23$\pm$0.62 & 68.85$\pm$0.32 & 62.92$\pm$1.02 & 71.69$\pm$0.51 &  \textbf{78.34$\pm$0.15} \\															
				SA Hart      & 69.59$\pm$0.08 & 61.69$\pm$0.44 & 70.91$\pm$0.42 & 68.7$\pm$0.46 & 70.05$\pm$0.51 & \textbf{73.92$\pm$0.72} & 72.8$\pm$0.82\\
				Sonar         & 53.24$\pm$0.62 & 54.93$\pm$0.26 & 54.1$\pm$0.91 & 54.39$\pm$0.25 & 57.64$\pm$0.47 & 52.06$\pm$0.873 & \textbf{61.29$\pm$0.11}\\
				Statlog        & 42.87$\pm$0.62 & 52.35$\pm$0.79 & 54.88$\pm$0.528 & 55.77$\pm$0.719 & 53.73$\pm$0.52 & 55.76$\pm$0.591 & \textbf{57.92$\pm$0.26}\\
				USPS         & 62.67$\pm$0.13 & 59.72$\pm$0.62 & 63.91$\pm$0.94 & 65.21$\pm$0.69 & 68.73$\pm$0.66 & 65.38$\pm$1.02 & \textbf{70.37$\pm$0.01}\\
				Wine           & 73.09$\pm$1.38 & 83.81$\pm$0.41 & 64.6$\pm$0.231 & 71.34$\pm$0.542 & 88.46$\pm$0.71 & 87.34$\pm$0.24 & \textbf{90.44$\pm$0.02}\\
				Yeast          & 32.96$\pm$0.71 & 30.49$\pm$0.63 & 31.06$\pm$0.245 & 32.76$\pm$0.268 & 35.19$\pm$0.57 & 28.12$\pm$0.462 & \textbf{36.92$\pm$0.81}\\
				\hline
			\end{tabular}
		\end{small}
	\end{center}
	\vskip -0.1in
\end{table*}
\subsection{Performance analysis}
In this section the performance (accuracy metric \cite{Alizadeh15}) of proposed method will be analyzed. In other words, the final clustering performance was evaluated by re-labeling between obtained clusters and the ground truth labels and then counting the percentage of correctly classified samples \cite{Alizadeh15}. The results of the proposed method are compared with individual algorithms Spectral clustering\cite{Ng01} and MLE \cite{Chen14}, as well as APMM \cite{Alizadeh14}, WOCCE \cite{Alizadeh15}, SMI \cite{Romano14}, and BGCM \cite{Gao13} which are state-of-the-art cluster ensemble (selection) methods. The main reason for comparing the proposed method with Spectral clustering is to show the effect of TKSC framework on the performance of the final results. Furthermore, as a new alternative in the graph based clustering methods, the empirical results of WSCE are compared with the MLE and BGCM methods. This paper uses the unsupervised version of BGCM method (with the null set of supervision information). For representing the effect of Normalized Modularity on the performance of the final results, it compares with three state-of-the-art metrics in diversity evaluation (A3, APMM and SMI), which are based on Shannon’s entropy. This paper doesn't use optional feature selection in this section ($d = 0$). The experimental results are given in Table II. In this table, the best result which is achieved for each data set is highlighted in bold. As depicted in this table, although individual clustering algorithms (Spectral and MLE) have shown acceptable performance in some data sets, they cannot recognize true patterns in all of them. As mentioned earlier in this paper, in order to solve the clustering problem, each individual algorithm considers a special perspective of a data set which is based on its objective function. The achieved results of individual clustering algorithms, which are depicted in Table II are good evidence for this claim. Furthermore, the results generated by APMM, SMI, and WOCCE show the effect of the aggregation method on improving accuracy in the final results. According to Table II, BGCM and the proposed algorithm (WSCE) have generated better results in comparison with other individual and ensemble algorithms. Even though the proposed method was outperformed by a number of algorithms in four data sets (Iris, SA Hart, and ADNI-MRI-C1/C2), the majority of the results demonstrate the superior accuracy of the proposed method in comparison with other algorithms. In addition, the difference between the performance of proposed method and the best result in those four data sets is lower that 2\%.
\begin{figure*}[ht]
	\vskip 0.2in
	\begin{center}
		\begin{minipage}{0.48\linewidth}
			\includegraphics[width=0.85\textwidth]{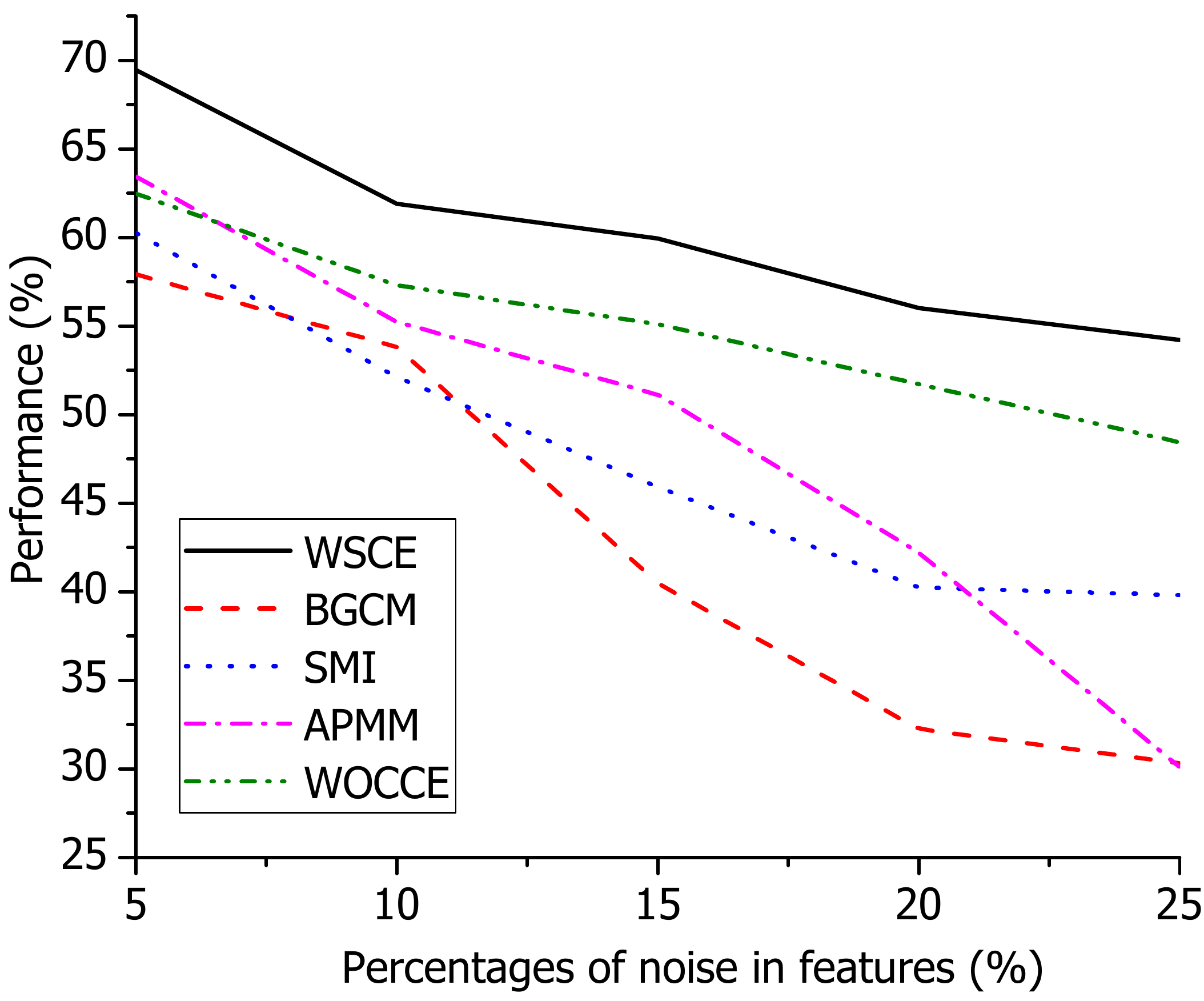}\\
			\centering (a) Arcene
		\end{minipage}
		\begin{minipage}{0.48\linewidth}
			\includegraphics[width=0.85\textwidth]{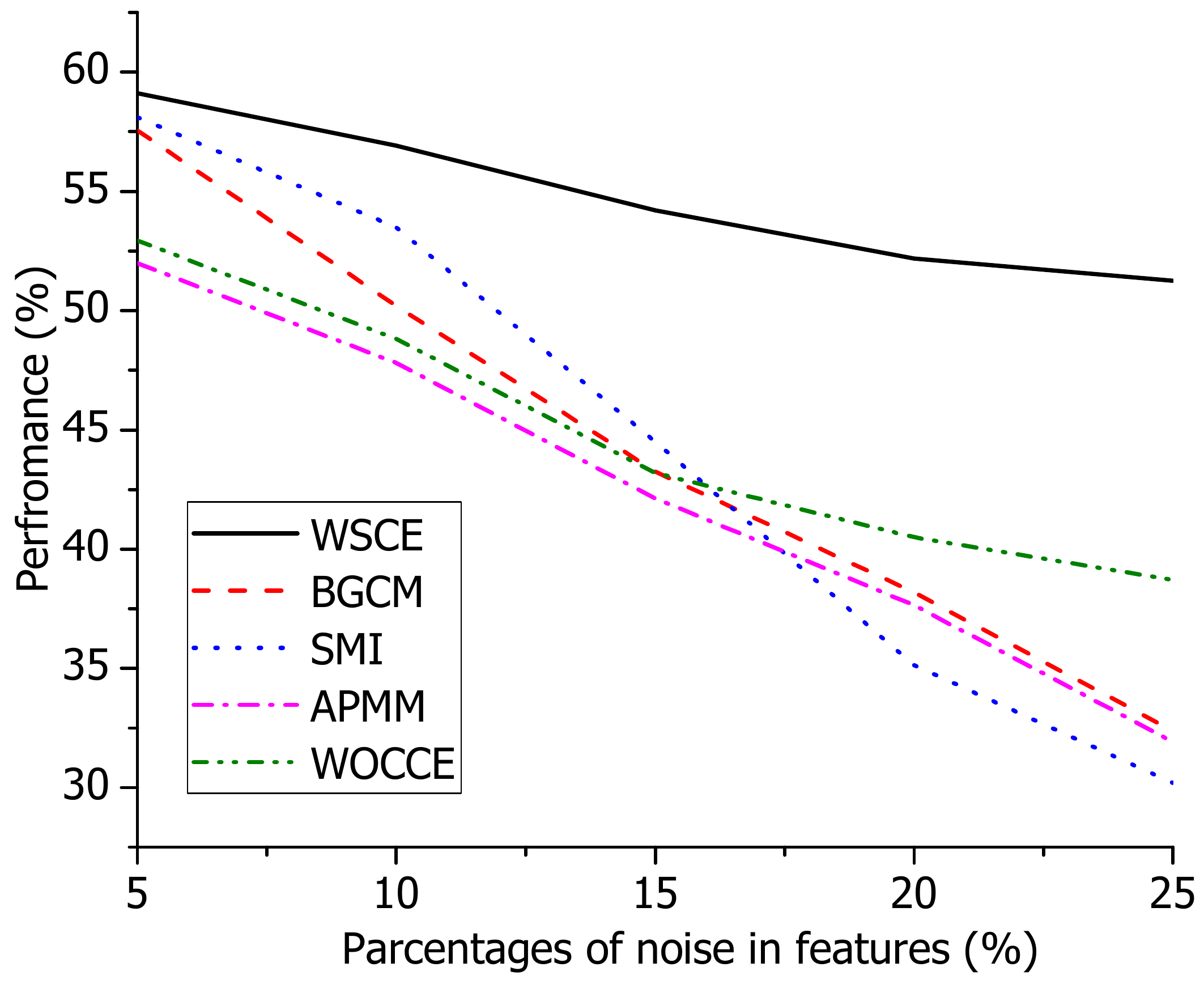}\\
			\centering (b) CNAE-9
		\end{minipage}
		\caption{The effect of noisy data sets on the performance.}
		\label{Fig2}
	\end{center}
	\vskip -0.2in
\end{figure*}
\begin{figure*}[ht]
	\vskip 0.2in
	\begin{center}
		\begin{minipage}{0.48\linewidth}
			\includegraphics[width=0.85\textwidth]{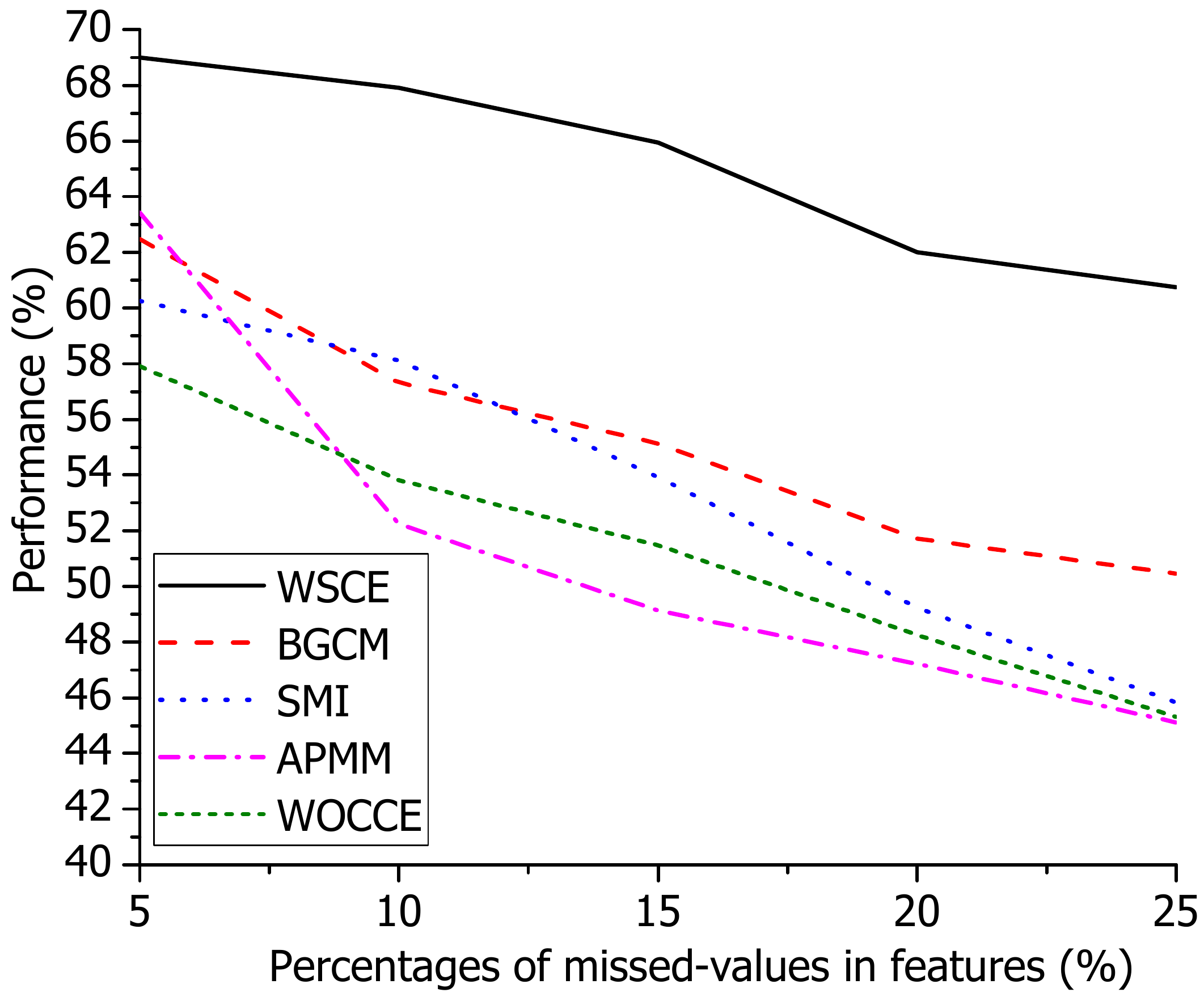}\\
			\centering (a) Arcene
		\end{minipage}
		\begin{minipage}{0.48\linewidth}
			\includegraphics[width=0.85\textwidth]{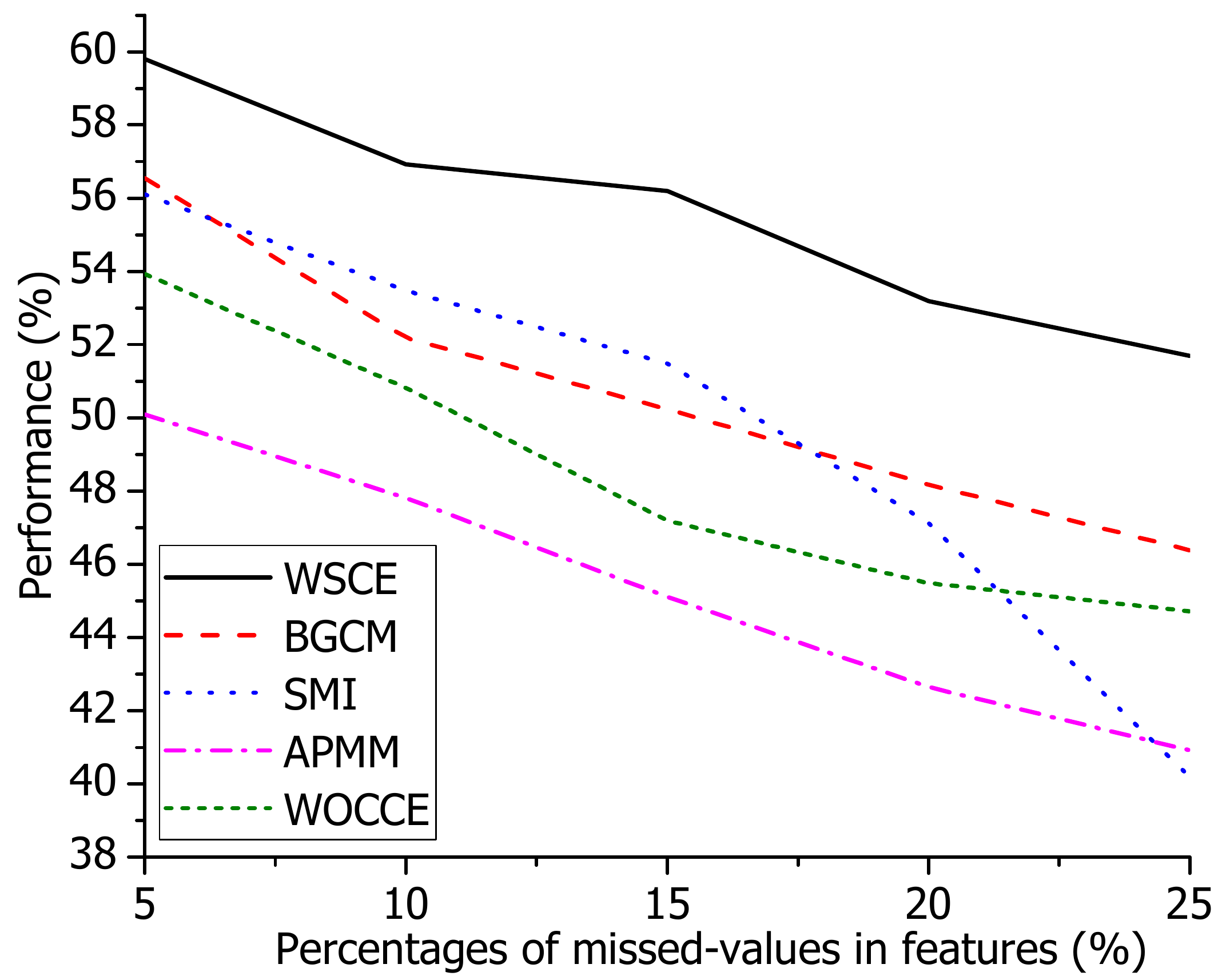}\\
			\centering (b) CNAE-9
		\end{minipage}
		\caption{The effect of missed-values on the performance.}
		\label{Fig3}
	\end{center}
	\vskip -0.2in
\end{figure*}
\subsection{Noise and missed-values analysis}
The effect of noise and missed-values on the performance of clustering algorithms will be discussed in this section. The optional feature selection for the proposed method doesn't use in this section (d = 0). In Figure 2, the effect of noise in the features of data sets will be analyzed on the performance of proposed method. This figure represents the performance of the WSCE, WOCCE, BGCM, SMI, and APMM on the noisy data sets. In this experiment, some features of Arcene and CNAE-9 data sets are randomly changed. This figure shows that proposed method generates more stable results because the Normalized Modularity provides a robust diversity evaluation for selecting most stable individual results. As mentioned before, Shannon's entropy uses the logarithm of probability of individual results for evaluating the diversity but there is no mathematical prove that all real-world data sets have logarithmic behavior. This experiment is the best evidence for this claim. Figure 3 demonstrates the analysis for the effect of missed-values in the data sets on the performance of clustering algorithms. This figure illustrates the performance of the WSCE, WOCCE, BGCM, SMI, and APMM on the data sets with missed-values. In this experiment, some values of attributes of Arcene and CNAE-9 data sets are randomly missed (set null). As you can see in this Figure, the proposed method and BGCM generate more stable results. This is a new advantage of our proposed method in comparison other non-graph based methods. Since, our proposed method uses the TKSC algorithms for generating Partitional and Modular results, it can significantly handle the miss values. In other words, as a local error in the individual results, a missed-value just can destroy an edge in our Modular result, which can be recognized by comparing Modular result with Partitional result in the diversity evaluation by using the NM metric. That is another reason for exploiting the proposed framework in the clustering problems. 
\subsection{Parameter analysis}
In this section the performance of the proposed method will be analyzed by using the optional features selection ($d$ parameter). This paper employs various data sets, i.e. two low dimension data sets (Wine, Glass), two high-dimension data sets (20 Newsgroups, Arcene), and two middle-dimension and also image based data sets (USPS, ADNI) for analyzing the performance of proposed method. Figure 4 illustrates the relationship between the performance of the proposed method based on the percentage of selected features in different data sets. The vertical axis refers to the performance while the horizontal axis refers to the percentage of selected feature in each data set. As you can see in this figure, the optional feature selection can significantly increase the performance of final results on high-dimensional data sets; and also it can dramatically decrease the performance on low-dimensional data sets. Further, it is not more effective on the middle-dimension data sets. This paper offers that the optional features selection will be used only for high-dimensional data sets for handling features-sparsity. 
\begin{figure}[!t]
	\centering
	\includegraphics[width=0.85\columnwidth]{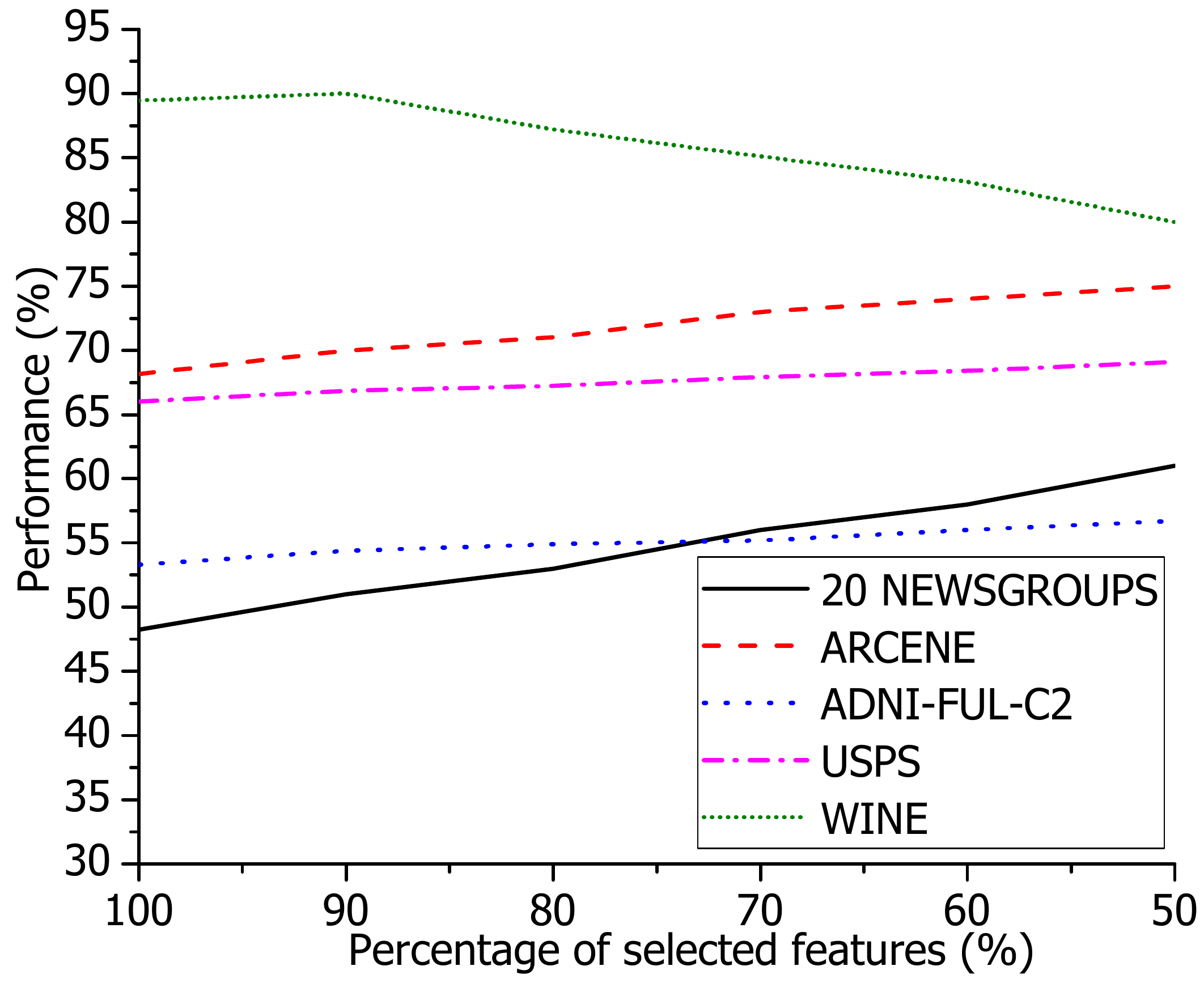}
	\caption{The effect of optional features selection on the performance of proposed method.}
	\label{Fig4}
\end{figure}
\section{Conclusion}
There are two challenges in Cluster Ensemble Selection (CES); i.e. proposing a robust consensus metric(s) for diversity evaluation and estimating optimum parameters in the thresholding procedure for selecting the evaluated results. This paper introduces a novel solution for solving mentioned challenges. By employing some concepts from community detection arena and graph based clustering, this paper proposes a novel framework for clustering problems, which is called Weighted Spectral Cluster Ensemble (WSCE). Under this framework, a new version of spectral clustering, which is called Two Kernels Spectral Clustering (TKSC), is used for generating graphs based individual clustering results; i.e. Partitional result and Modular result. Instead of entropy based methods in the traditional CES, this paper introduces Normalized Modularity (NM), which is a redefined version of modularity in the community detection arena for general clustering problems. The NM is used on the transformed graph representation of individual clustering results for providing an effective diversity estimation. Moreover, this paper introduces a new solution for combining the evaluated individual clustering results without the procedure of thresholding, which is called Weighted Evidence Accumulation Clustering (WEAC). While the weight of each individual result in WEAC can be estimated with different metrics, the NM was used in this paper. To validate the effectiveness of the proposed approach, an extensive experimental study is performed by comparing with individual clustering methods as well as cluster ensemble (selection) methods on a large number of data sets. Results clearly show the superiority of our approach on both normal data sets and those with noise or missing values.  In the future, we plan to develop a new version of normalized modularity for estimating the diversity of Partitional results, directly. 
\section*{Acknowledgment}
We thank Dr. Sheng-Jun Huang for his helpful suggestions, and the anonymous reviewers for comments. This work was supported in part by the National Natural Science Foundation of China (61422204 and 61473149), Jiangsu Natural Science Foundation for Distinguished Young Scholar (BK20130034) and NUAA Fundamental Research Funds (NE2013105). 
\bibliographystyle{IEEEtran}
\bibliography{WSCE_Edited}

\begin{thebibliography}{10}
\providecommand{\url}[1]{#1}
\csname url@samestyle\endcsname
\providecommand{\newblock}{\relax}
\providecommand{\bibinfo}[2]{#2}
\providecommand{\BIBentrySTDinterwordspacing}{\spaceskip=0pt\relax}
\providecommand{\BIBentryALTinterwordstretchfactor}{4}
\providecommand{\BIBentryALTinterwordspacing}{\spaceskip=\fontdimen2\font plus
\BIBentryALTinterwordstretchfactor\fontdimen3\font minus
  \fontdimen4\font\relax}
\providecommand{\BIBforeignlanguage}[2]{{%
\expandafter\ifx\csname l@#1\endcsname\relax
\typeout{** WARNING: IEEEtran.bst: No hyphenation pattern has been}%
\typeout{** loaded for the language `#1'. Using the pattern for}%
\typeout{** the default language instead.}%
\else
\language=\csname l@#1\endcsname
\fi
#2}}
\providecommand{\BIBdecl}{\relax}
\BIBdecl

\bibitem{Strehl02}
A.~Strehl and J.~Ghosh, ``Cluster ensembles - a knowledge reuse framework for
  combining multiple partitions,'' \emph{Journal of Machine Learning Research},
  vol.~3, pp. 583--617, 2002.

\bibitem{Fred08}
A.~Fred and A.~Lourenco, ``Cluster ensemble methods: from single clusterings to
  combined solutions,'' \emph{Computer Intelligence}, vol. 126, pp. 3--30,
  2008.

\bibitem{Fern08}
X.~Fern and W.~Lin, ``Cluster ensemble selection,'' in \emph{SIAM International
  Conference on Data Mining (SDM'08)}, 24-26 April 2008, pp. 128--141.

\bibitem{Azimi09}
J.~Azimi and X.~Fern, ``Adaptive cluster ensemble selection,'' in \emph{21th
  International joint conference on artificial intelligence (IJCAI-09)}, 11-17
  July 2009, pp. 992--997.

\bibitem{Alizadeh14}
H.~Alizadeh, B.~Minaei-Bidgoli, and H.~Parvin, ``Cluster ensemble selection
  based on a new cluster stability measure,'' \emph{Intelligence Data Analysis
  (IDA)}, vol.~18, no.~3, pp. 389--40, 2014.

\bibitem{Jia12}
J.~Jia, X.~Xiao, and B.~Liu, ``Similarity-based spectral clustering ensemble
  selection,'' in \emph{9th International Conference on Fuzzy Systems and
  Knowledge Discovery (FSKD)}, 29-31 May 2012, pp. 1071--1074.

\bibitem{Yousefnezhad13}
M.~Yousefnezhad, H.~Alizadeh, and B.~Minaei-Bidgoli, ``New cluster ensemble
  selection method based on diversity and independent metrics,'' in \emph{5th
  Conference on Information and Knowledge Technology (IKT'13)}, 22-24 May 2013.

\bibitem{Alizadeh15}
H.~Alizadeh, M.~Yousefnezhad, and B.~Minaei-Bidgoli, ``Wisdom of crowds cluster
  ensemble,'' \emph{Intelligent Data Analysis (IDA)}, vol.~19, no.~3, 2015.

\bibitem{Clauset04}
A.~Clauset, M.~Newman, and C.~Moore, ``Finding community structure in very
  large networks,'' \emph{Physical Review E}, vol.~70, no. 066111, 2004.

\bibitem{Newman06}
M.~E.~J. Newman, ``Modularity and community structure in networks,''
  \emph{Proceedings of the National Academy of Sciences of the United States of
  America}, vol. 103, no.~23, pp. 8577--8696, 2006.

\bibitem{Zhang14}
X.~Zhang, L.~Zhao, L.~Zong, and X.~Liu, ``Multi-view clustering via
  multi-manifold regularized nonnegative matrix factorization,'' in \emph{IEEE
  International Conference on Data Mining series (ICDM'14)}, 15--17 December
  2014.

\bibitem{Fred05}
A.~Fred and A.~K. Jain, ``Combining multiple clusterings using evidence
  accumulation,'' \emph{IEEE Transaction on Pattern Analysis and Machine
  Intelligence}, vol.~27, pp. 835--850, 2005.

\bibitem{Romano14}
S.~Romano, J.~Bailey, N.~X. Vinh, and K.~Verspoor, ``Standardized mutual
  information for clustering comparisons: One step further in adjustment for
  chance,'' in \emph{31st International Conference on Machine Learning
  (ICML14)}, 21-26 June 2014, pp. 1143--1151.

\bibitem{Kuo14}
C.-T. Kuo, P.~Walker, O.~Carmichael, and I.~Davidson, ``Spectral clustering for
  medical imaging,'' in \emph{IEEE International Conference on Data Mining
  series (ICDM'14)}, 15--17 December 2014.

\bibitem{Chen14}
Y.~Chen, S.~H. Lim, and H.~Xu, ``Weighted graph clustering with non-uniform
  uncertaintiese,'' in \emph{31st International Conference on Machine Learning
  (ICML14)}, 21-26 June 2014, pp. 1566--1574.

\bibitem{Gao13}
J.~Gao, F.~Liang, W.~Fan, Y.~Sun, and J.~Han, ``A graph-based consensus
  maximization approach for combining multiple supervised and unsupervised
  models,'' \emph{IEEE Transactions on Knowledge and Data Engineering},
  vol.~25, no.~1, pp. 15--2, 2013.

\bibitem{Ng01}
A.~Ng, M.~Jordan, and Y.~Weiss, ``On spectral clustering: Analysis and an
  algorithm,'' in \emph{Advances in Neural Information Processing Systems 14
  (NIPS'01)}, 2001, pp. 849--856.

\bibitem{Yan09}
D.~Yan, L.~Huang, and M.~I. Jordan, ``Fast approximate spectral clustering,''
  in \emph{15th ACM Conference on Knowledge Discovery and Data Mining
  (SIGKDD)}, 2009.

\bibitem{USPS}
\BIBentryALTinterwordspacing
S.~Roweis. (1998) The world-famous courant institute of mathematical sciences,
  computer science department, new york university. [Online]. Available:
  \url{http://cs.nyu.edu/∼roweis/data.html}
\BIBentrySTDinterwordspacing

\bibitem{Zu14}
C.~Zu and D.~Zhang, ``Label-alignment-based multi-task feature selection for
  multimodal classification of brain disease,'' in \emph{4th NIPS Workshop on
  Machine Learning and Interpretation in Neuroimaging (MLINI'14)}, 13 December
  2014.

\bibitem{UCI}
\BIBentryALTinterwordspacing
C.~B. D.~J. Newman, S.~Hettich, and C.~Merz. (1998) Uci repository of machine
  learning databases. [Online]. Available:
  \url{http://www.ics.uci.edu/mlearn/MLSummary.html}
\BIBentrySTDinterwordspacing

\end{thebibliography}
\end{document}